\documentclass{article}
\usepackage{spconf,amsmath,graphicx}
\usepackage{caption}
\usepackage{subcaption}
\usepackage{multirow}
\usepackage{tikz}
\usepackage{eso-pic}

\def\checkmark{\tikz\fill[scale=0.4](0,.35) -- (.25,0) -- (1,.7) -- (.25,.15) -- cycle;} 

\usepackage[absolute,showboxes]{textpos}
\TPMargin{5pt}
\newcommand{\copyrightstatement}{
    \begin{textblock}{0.84}(0.08,0.93)    
         \noindent
         \footnotesize
         \copyright 2020 IEEE. Personal use of this material is permitted. Permission from IEEE must be obtained for all other uses, in any current or future media, including reprinting/republishing this material for advertising or promotional purposes, creating new collective works, for resale or redistribution to servers or lists, or reuse of any copyrighted component of this work in other works.
    \end{textblock}
}

\title{SEGMENTING UNSEEN INDUSTRIAL COMPONENTS IN A HEAVY CLUTTER USING RGB-D FUSION AND SYNTHETIC DATA}
\name{Seunghyeok Back, Jongwon Kim, Raeyoung Kang, Seungjun Choi, Kyoobin Lee \thanks{This work was supported by Institute for Information \& Communications Technology Promotion(IITP) grant funded by Korea goverment(MSIT) (No.2019-0-01335, Development of AI technology to generate and validate the task plan for assembling furniture in the real and virtual environment by understanding the unstructured multi-modal information from the assembly manual.}}
\address{Gwangju Institute of Science and Technology (GIST), Republic of Korea}
\begin{document}
%

\maketitle
\copyrightstatement

\begin{abstract}
Segmentation of unseen industrial parts is essential for autonomous industrial systems. However, industrial components are texture-less, reflective, and often found in cluttered and unstructured environments with heavy occlusion, which makes it more challenging to deal with unseen objects. To tackle this problem, we present a synthetic data generation pipeline that randomizes textures via domain randomization to focus on the shape information. In addition, we propose an RGB-D Fusion Mask R-CNN with a confidence map estimator, which exploits reliable depth information in multiple feature levels. We transferred the trained model to real-world scenarios and evaluated its performance by making comparisons with baselines and ablation studies. We demonstrate that our methods, which use only synthetic data, could be effective solutions for unseen industrial components segmentation.
\end{abstract}

\section{Introduction}
\label{sec:intro}
In the industrial domain, the detection of unseen objects is a key necessity for various industrial automation and robotics applications, such as warehouse bin-picking, sorting, and robotic assembly. Recent advances in deep learning have enabled pixelwise instance segmentation of objects in the broad domain, but generalization performance of deep learning models relies heavily on the amount and quality of training dataset. Thus, building a large-scale, high-quality dataset is crucial for robust and industry-applicable performance. However, repeating the data collecting and labeling procedure manually is costly and often infeasible. Utilizing a large amount of synthetic data for model training can be an inexpensive and effective solution. Through simulations, images and corresponding labels are auto-generated. Additionally, virtual environments can be easily altered, which can offer the flexibility required to deal with fast production cycles and improve the robustness of the model. Recently, the concept of category-agnostic instance segmentation \cite{danielczuk2019segmenting,xie2019best} has been proposed to detect unseen segment objects efficiently. It has been shown that exploiting deep learning models such as Mask R-CNN \cite{he2017mask} with a large amount of synthetic data for segmentation is useful to generalize unseen objects. Though the results are promising, unseen object instance segmentation for industrial applications has not been studied in detail.

\begin{figure}[t!]
\centering
\begin{subfigure}[t]{0.49\columnwidth}
    \centering
    \includegraphics[width=\textwidth]{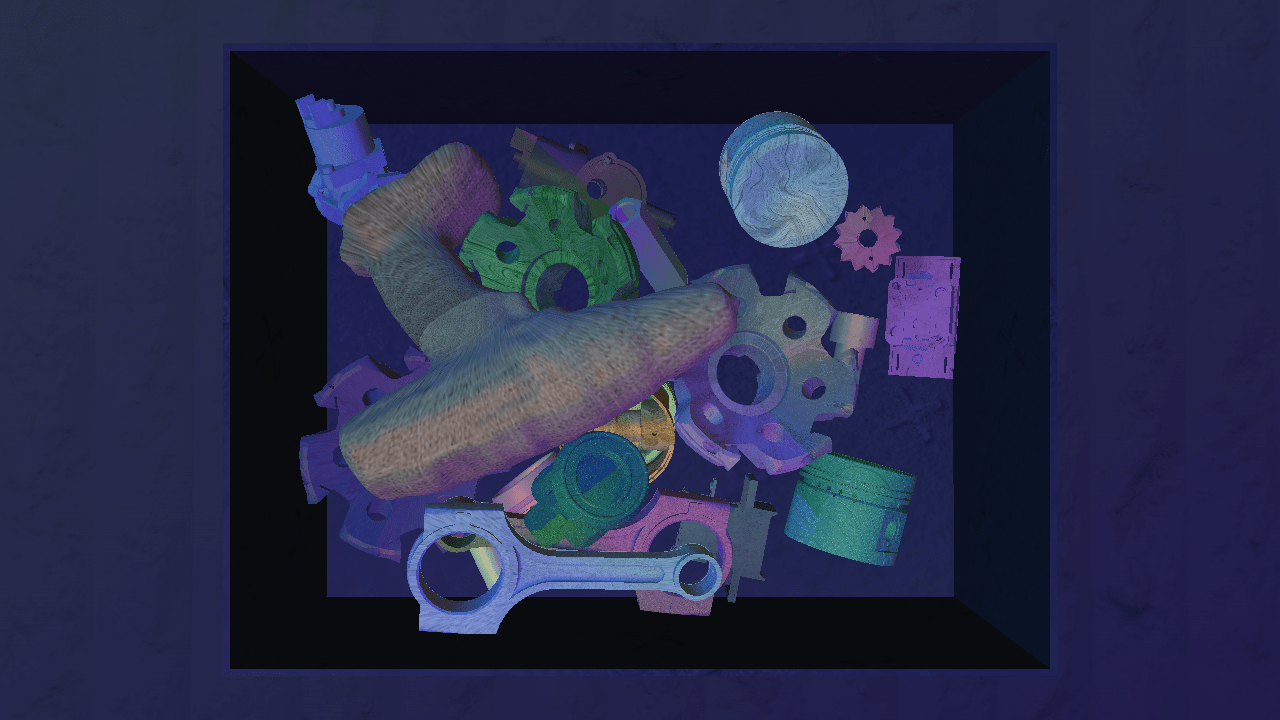}
    \caption{RGB image}
\end{subfigure}
\begin{subfigure}[t]{0.49\columnwidth}
    \centering
    \includegraphics[width=\textwidth]{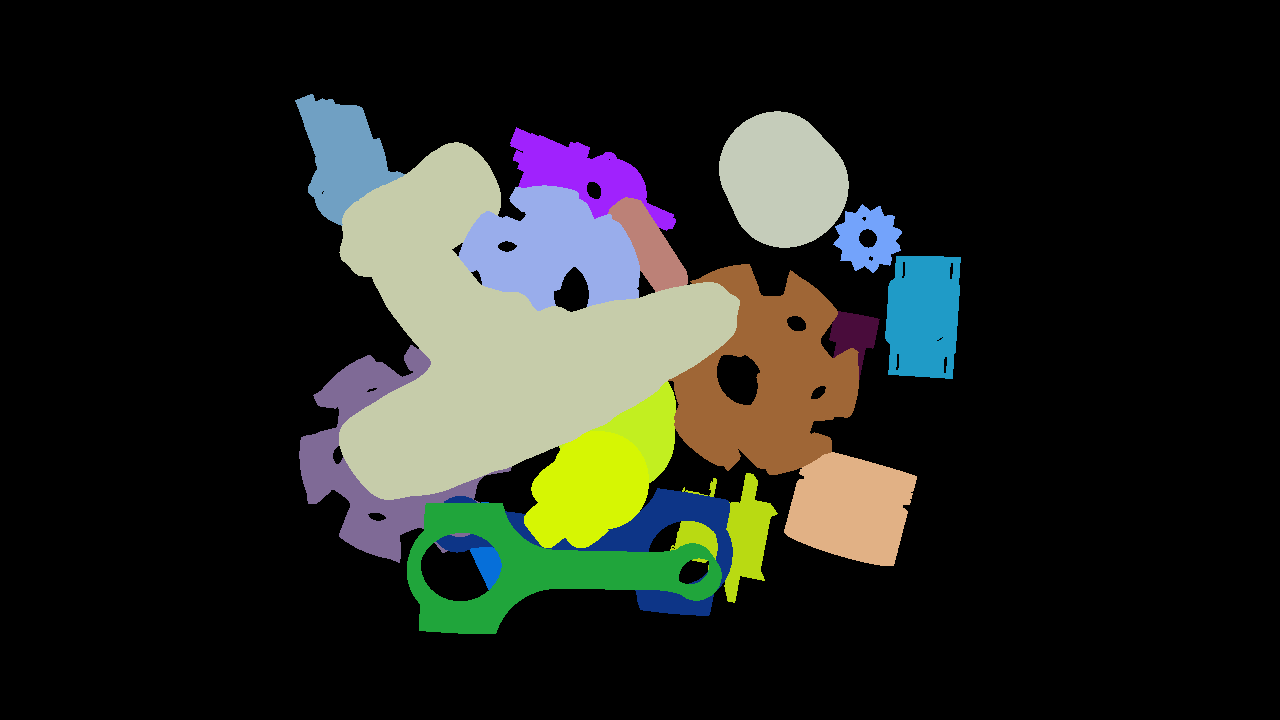}
    \caption{Segmentation mask}
\end{subfigure}
\begin{subfigure}[t]{0.49\columnwidth}
    \centering
    \includegraphics[width=\textwidth]{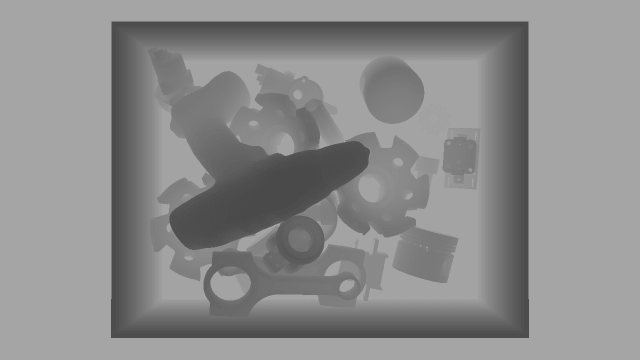}
    \caption{Depth image (initial)}
\end{subfigure}
\begin{subfigure}[t]{0.49\columnwidth}
    \centering
    \includegraphics[width=\textwidth]{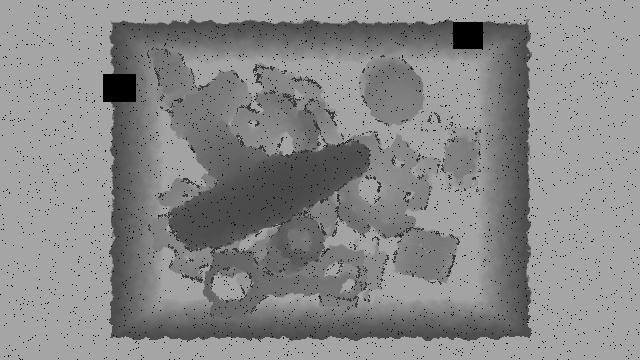}
    \caption{Depth image (augmented)}
\end{subfigure}
\caption{Examples of synthetic data generated with our proposed method.}\label{fig:synthetic1}
\end{figure}

Segmenting unseen industrial components is quite challenging due to its unique properties. Many industrial components are texture-less, metallic, and shiny so that their visual appearances vary based on light and object configuration \cite{rodrigues20126d, hodan2017t}, while high reflectivity causes missing values and noise in depth maps. They are also often placed in the clutter with heavy occlusion in an unstructured environment while the shape and size of industrial components vary. Besides, only a few industry-relevant public datasets are available, and these datasets lack variation in the number of images and objects. For example, the T-less dataset \cite{hodan2017t} only includes texture-less objects while ITODD \cite{drost2017introducing} only offers 800 scenes of industrial objects. Though \cite{kleeberger2019large} offers a vast amount of image pairs in a bin clutter environment, it includes only 10 industrial objects. 

For metallic component detection, Lee et al. proposed a method to generate photorealistic synthetic training data for the segmentation of wrenches in the real world \cite{lee2019automatic}. However, they do not concern detecting novel objects but focus on the segmentation of only a single wrench, and fine-tuning using real data should be performed. On the other hand, domain randomization \cite{tobin2017domain} can be promising since it can be used for the sim-to-real transfer of the deep learning model by utilizing only synthetic data, while domain adaptation \cite{sankaranarayanan2018learning} requires real data for feature learning. Rendering highly photorealistic synthetic data \cite{hodavn2019photorealistic} can be an alternative to reduce the dependency on real-data, but this requires carefully designed simulations and high computation costs, which offers low flexibility and is not suitable for industrial purposes. 

In this paper, we present a synthetic data generation method that simulates an industrial bin environment with high clutter and occlusion by randomizing textures and adding depth noise. We propose an RGB-D Fusion Mask R-CNN with a confidence module that exploits RGB and depth more effectively. We trained our model using a synthetic dataset and evaluated its performance by conducting a varying input modality ablation study. Through experiments, we demonstrate that the proposed method could be applied to segment unseen industrial components, and fusing RGB and depth with a confidence map estimation can improve industrial component segmentation performance.

\section{Method}

For the detection of unseen industrial objects, we generated a large-scale dataset that consists of 35,000 RGB-Depth-Mask pairs in a bin clutter environment. Then, we trained the category-agnostic instance segmentation model on a synthetic dataset and transferred the model to the real world without fine-tuning using real-world data. To exploit RGB and depth inputs together, we adopted a confidence module to fuse RGB and depth in Mask R-CNN. We demonstrate that our trained model can segment unseen objects in the real world with non-realistic rendering.

\subsection{Synthetic Data Generation}

To segment industrial components within a clutter, the model should focus on an object’s visual shape while ignoring texture information due to an object’s texture-less, reflective properties. By randomly texturing the objects, a trait adopted from domain randomization \cite{tobin2017domain}, we can encourage the model to learn domain-relevant features, such as the shape and edges, while reducing computational costs and human efforts required to optimize simulation environments.

\begin{figure}[t]
\centering
\begin{subfigure}[t]{0.50\columnwidth}
    \centering
    \includegraphics[width=\textwidth]{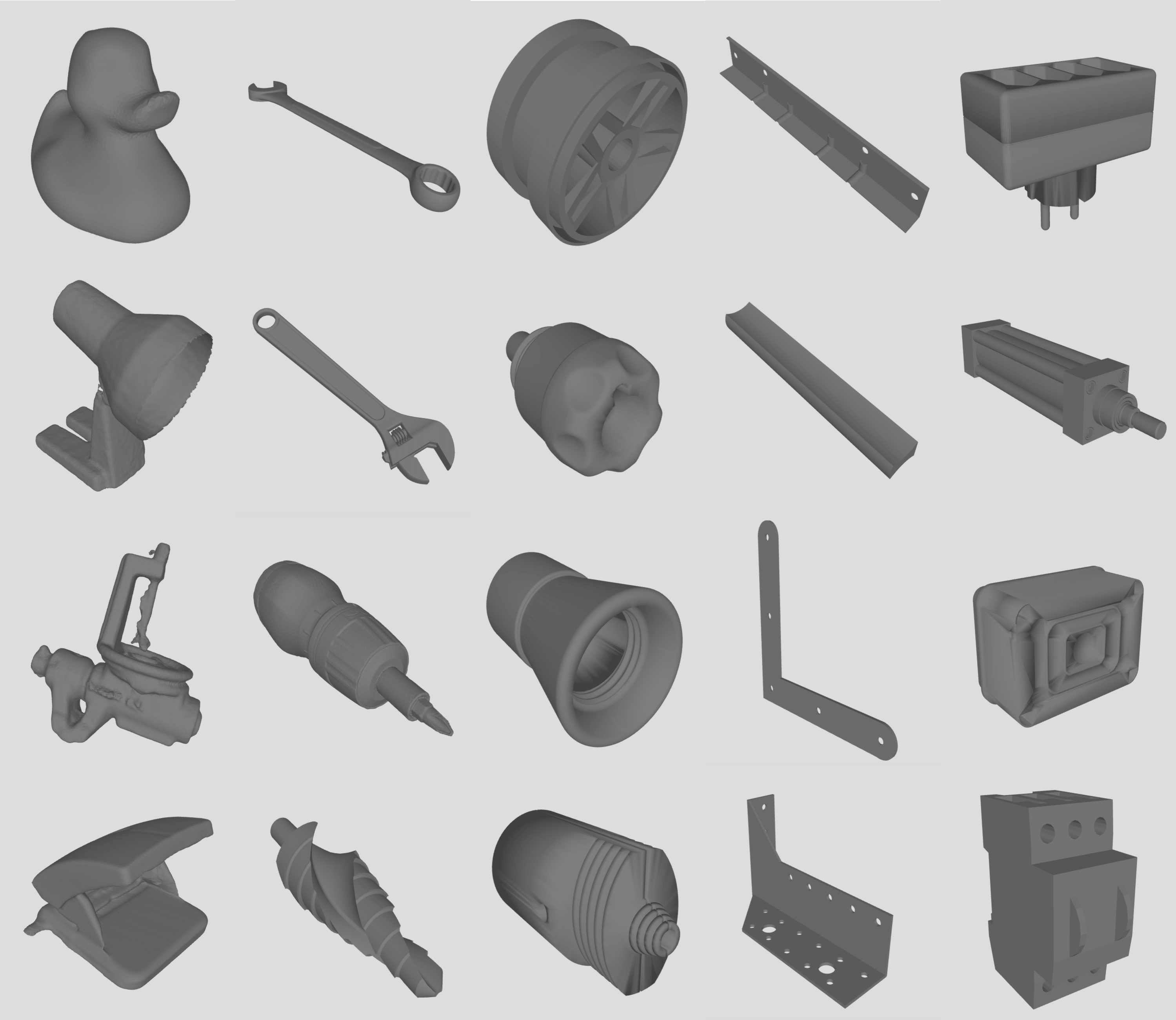}
\end{subfigure}
\begin{subfigure}[t]{0.432\columnwidth}
    \centering
    \includegraphics[width=\textwidth]{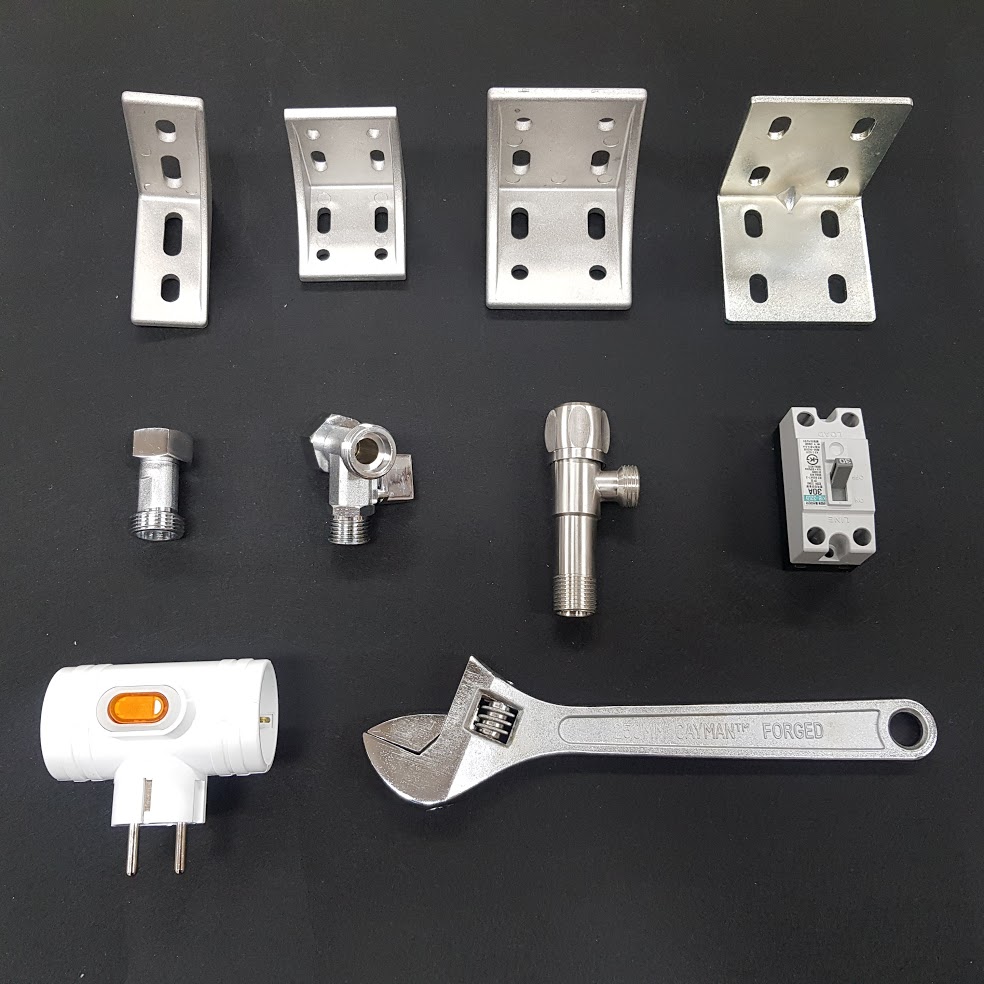}
\end{subfigure}
\caption{Left: Exemplary objects used for synthetic data generation, Right: Industrial objects for real data.}
\label{fig:exemplary_objects}
\end{figure}

We consider a typical industrial environment where multiple industrial components are placed in a bin with high clutter and occlusion. We collected a total of 149 3D CAD models from public datasets, including industrial objects \cite{drost2017introducing, DEKHTIAR2017:DMUNet, hodan2017t} and household objects \cite{hinterstoisser2012model, wohlhart2015learning} as shown in Fig. \ref{fig:exemplary_objects}.

\begin{figure}[t]
\centering
\begin{subfigure}[t]{0.47\columnwidth}
    \centering
    \includegraphics[width=\textwidth]{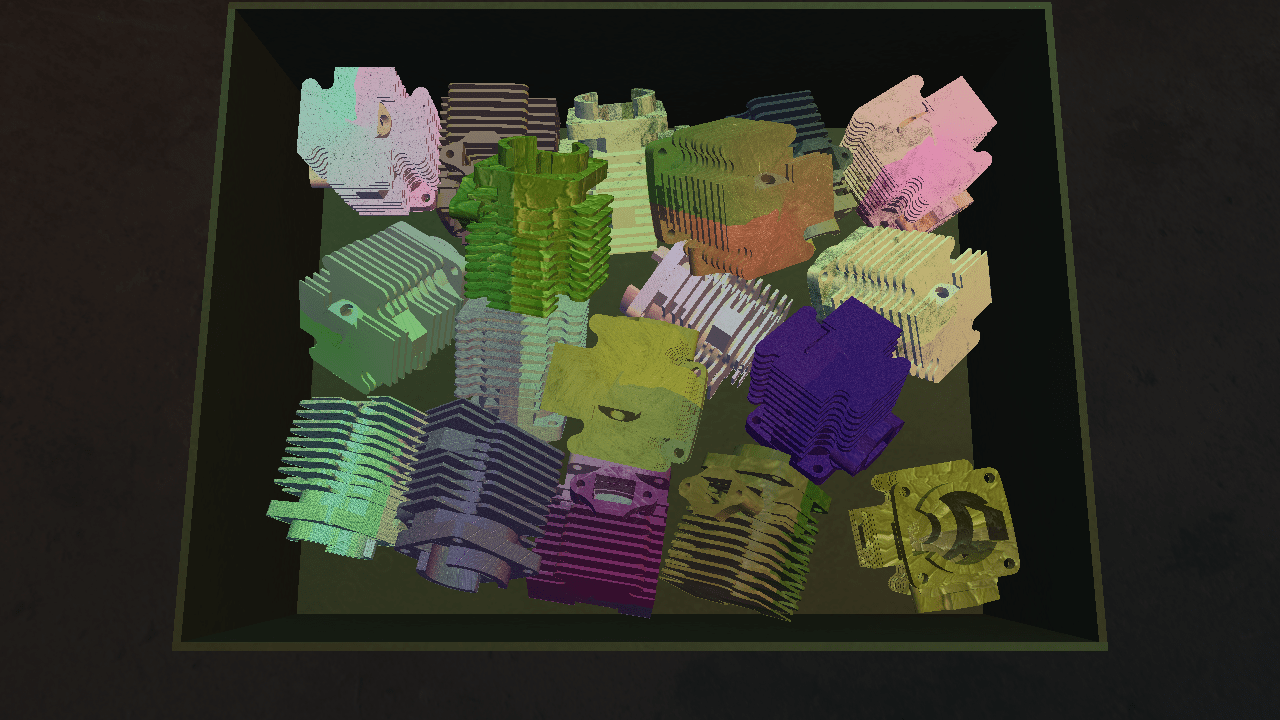}
\end{subfigure}
\begin{subfigure}[t]{0.47\columnwidth}
    \centering
    \includegraphics[width=\textwidth]{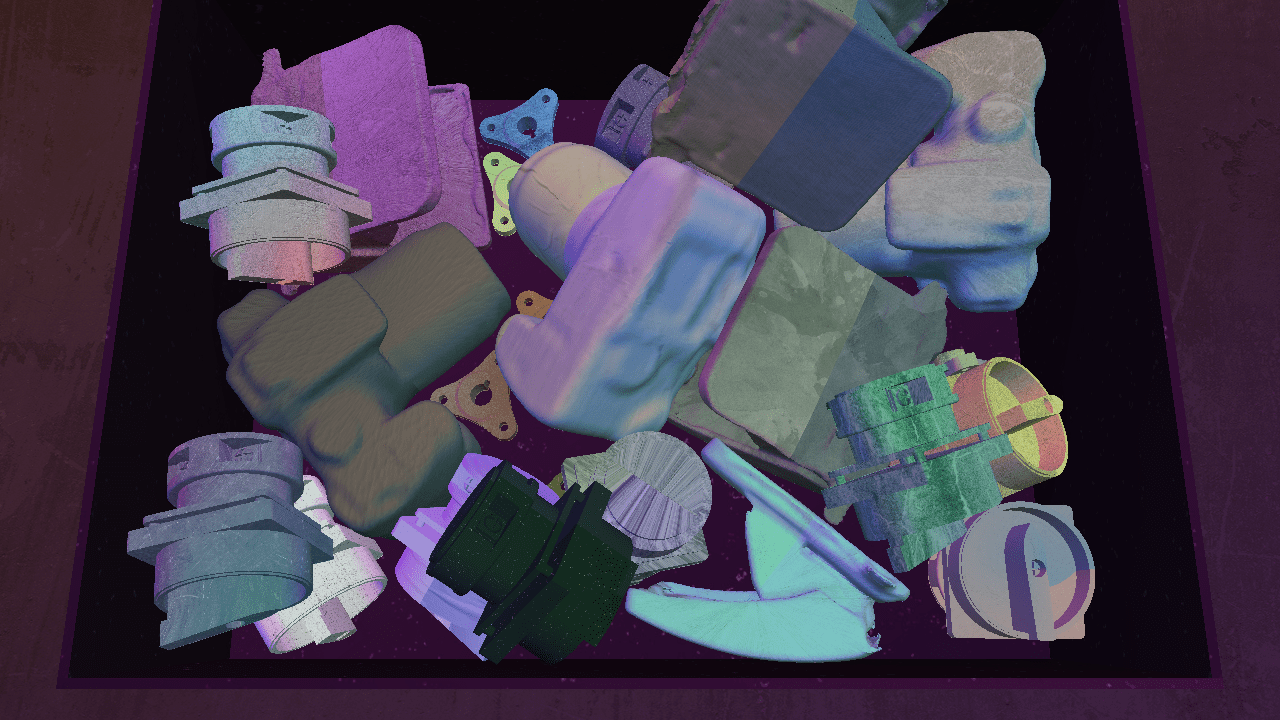}
\end{subfigure}
\begin{subfigure}[t]{0.47\columnwidth}
    \centering
    \includegraphics[width=\textwidth]{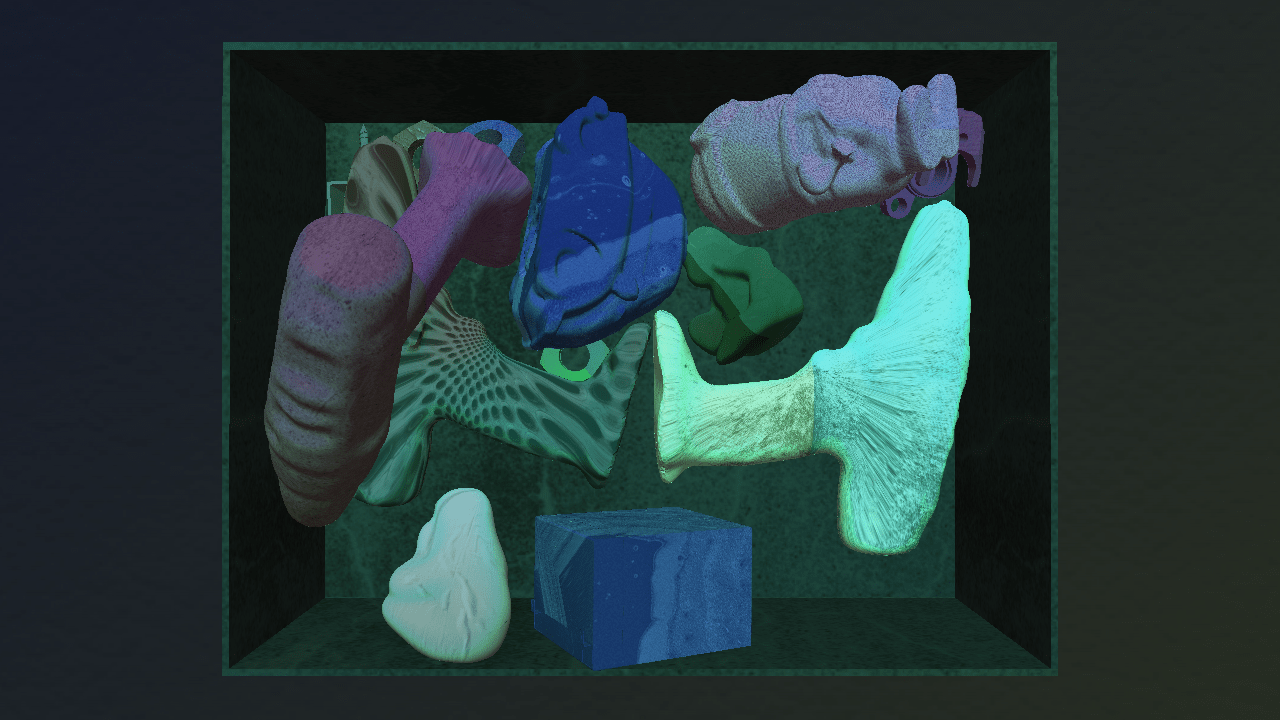}
\end{subfigure}
\begin{subfigure}[t]{0.47\columnwidth}
    \centering
    \includegraphics[width=\textwidth]{img/rgb4.png}
\end{subfigure}

\caption{Examples of domain randomized RGB images.}
\label{fig:synthetic2}
\end{figure}

\begin{figure*}[ht!]
\centering
  \includegraphics[width=0.95\textwidth]{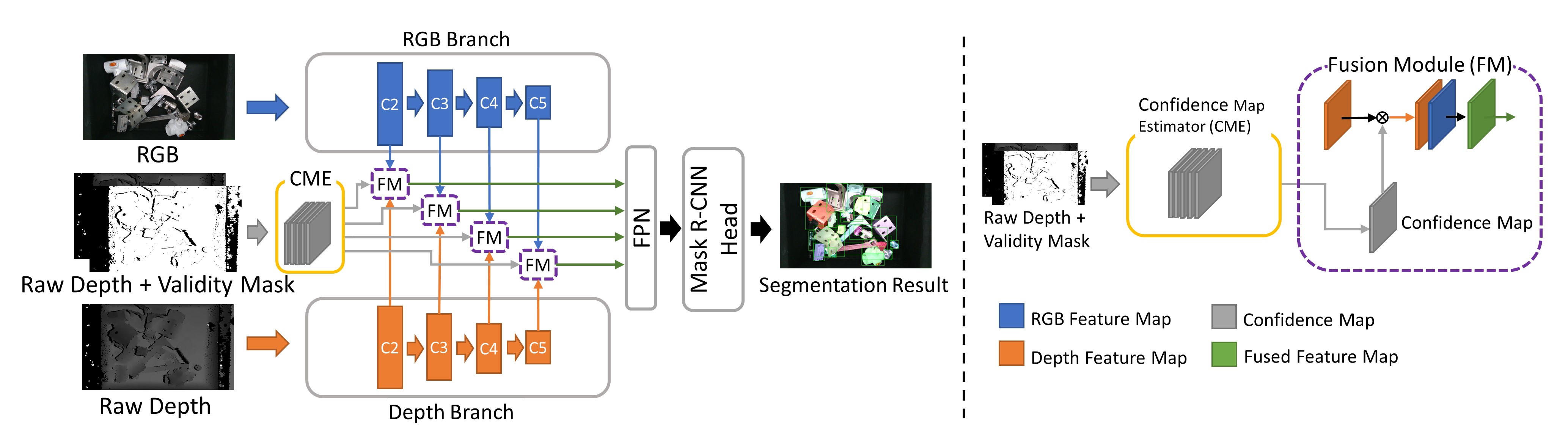}
  \caption{Proposed RGB-D fusion Mask R-CNN with confidence map estimator.}
  \label{fig:network_architecture}
\end{figure*}

Then, we configured a virtual bin-picking environment using V-REP \cite{rohmer2013v} to render images using physical simulations. For each configuration, we randomized the object’s properties, arrangement, and camera configurations. Specifically, 1-20 random objects were picked from the set and placed in the bin until the total number of objects became 40. For each object in the bin, we captured three RGB-Depth-Segmentation Mask pairs by varying the position, pose, texture, and color of each object. Additionally, the light intensity, camera’s distance, and orientation from the center of the bin were adjusted for each scene. Objects above the bin were removed, and the camera parameters were randomized to fit within the range detectable by the commercial RGB-D sensor Realsense D415. We repeated this procedure and obtained 30,000 training and 5000 validation images. Objects were also split into training and validation sets in the proportion 4:1 to prevent overlap.

While initial synthetic depth was complete and precise, depth maps in the real world are noisy and have missing values. To add realistic noises and sparsity to the synthetic depth map, we applied 3D Perlin noise \cite{zakharov2018keep} for random distortion. Then, we added random sparse edge noise using a segmentation mask and salt-and-pepper noise. Lastly, random erasing was performed to mimic the missing values in real depth maps. Examples of synthetic data are shown in Fig. \ref{fig:synthetic1} and \ref{fig:synthetic2}.

\subsection{RGB-D Fusion Mask R-CNN with Confidence Map Estimator}

When texture-less objects are piled within a clutter, using only color information would not be sufficient since occlusion boundaries become vague and dim in RGB images. In this situation, exploiting depth maps with RGB can improve segmentation performance since depths are likely to change sharply on the edge of objects. However, due to the reflective properties of metallic objects, raw depth maps are quite noisy and have missing values. Though raw or inpainted depth maps with hole-filling methods can be used as inputs for the model, this can lead to inaccurate results since the model assigns the same confidence to the whole depth map.

Fig. \ref{fig:network_architecture}  shows the overall architecture of our proposed network, which utilizes Mask R-CNN \cite{he2017mask} with an FPN backbone \cite{lin2017feature}. For RGB-D fusion, we adopted a fusion module (FM) and confidence map estimator (CME), which recalls the concept for a confidence map \cite{zeng2019deep} that was originally proposed for surface normal estimation tasks. In addition, we split the convolutional backbone with RGB and depth branches to handle different modalities with separate networks. ResNet-50 \cite{he2016deep}  was used as the backbone of each branch.

With the RGB and raw depth maps as inputs, our network extracts RGB and depth features from maps. Then, four different scales of RGB and depth features, (outputs of conv2, conv3, conv4 and conv5 [C2, C3, C4, C5]) in each branch are fused with the resized estimated confidence map in the fusion module. The confidence map estimator assigns pixel-wise reliability using the raw depth map and validity mask with five convolutional layers. The estimated confidence map is resized and multiplied with depth feature maps to apply spatial attention on the depth map. Through this operation, features with high confidence become larger values while ones with lower confidence become small. Then, RGB and depth feature maps are concatenated and converted into a fused feature map using the 1x1 convolution, which halves the number of channels. Finally, the fused feature maps are fed into the feature pyramid network (FPN) and the Mask R-CNN head and predict the bounding box and mask for the object and background.

\section{Experiments}
\label{sec:typestyle}
\begin{table}[t!]
\resizebox{\columnwidth}{!}{
\begin{tabular}{c|c|c|c|c|ccc}
\hline
\multirow{2}{*}{Method}     & \multirow{2}{*}{RGB} & \multicolumn{2}{c|}{Depth} & \multirow{2}{*}{\shortstack{RGB-D\\Fusion Strategy}} & \multicolumn{3}{c}{Metrics}                  \\ \cline{3-4} \cline{6-8} 
                            &                      & filled  & raw         &                         & AP$_{50}$            & AP            & AR            \\ \hline
\multirow{7}{*}{\shortstack{Mask \\R-CNN}} & \checkmark           &              &             &                         & 67.1          & 54.9          & 65.0          \\ \cline{2-5}
                            &                      & \checkmark   &             &                         & 61.2          & 52.6          & 62.4          \\ \cline{2-5}
                            &                      &              & \checkmark  &                         & 51.9          & 45.7          & 56.6          \\ \cline{2-5}
                            & \checkmark           & \checkmark   &             & Early Fusion          & 63.3          & 54.5          & 64.5          \\ \cline{2-5}
                            & \checkmark           &              & \checkmark  & Early Fusion          & 60.8          & 53.7          & 62.3          \\ \cline{2-5}
                            & \checkmark           & \checkmark   &             & Late Fusion             & 67.9          & 55.8          & 63.2          \\ \cline{2-5}
                            & \checkmark           &              & \checkmark  & Late Fusion              & 67.5          & 55.5          & 65.5          \\ \hline
Ours                        & \checkmark           &              & \checkmark  & Confidence Map          & \textbf{69.0} & \textbf{57.7} & \textbf{66.1} \\ \hline
\end{tabular}}
\caption{Performance comparison of the Mask R-CNN baselines and our network on real data.}
\label{table:results}
\end{table}

To evaluate the effectiveness of our synthetic data generation method, the category-agnostic segmentation performance of our model was examined on unseen real-world objects. Additionally, ablation studies were conducted to investigate the effects of input modality and fusion strategy in our proposed network. As evaluation metrics, average precision with IoU thresholds of 0.5 (AP$_{50}$), average precision (AP), and average recall (AR) using IoU thresholds from 0.50 to 0.95 with a 0.05 margin were used. The objects occluded by more than 80\% were excluded when computing the metrics.

For the performance evaluation of our proposed method in real-world scenarios, we collected 100 RGB-D image pairs using a commodity-level RGB-D sensor (Realsense D415) and labeled it manually. We captured RGB-D images in a bin environment by randomizing the objects and camera configurations within the randomization range performed in the simulations. As shown in Figure  \ref{fig:exemplary_objects}, objects used in the real dataset vary in shapes and size and has no direct relationship with the simulated objects. Depending on the number of objects in the scene, we define images with less than 15 objects as having a low occlusion and more than 15 objects as high occlusion while ensuring that the number of objects did not exceed 40. The real dataset includes 60 low-occlusion and 40 high-occlusion images.

\begin{figure}[t!]
\centering
\begin{subfigure}[t]{0.45\columnwidth}
    \centering
    \includegraphics[width=\textwidth]{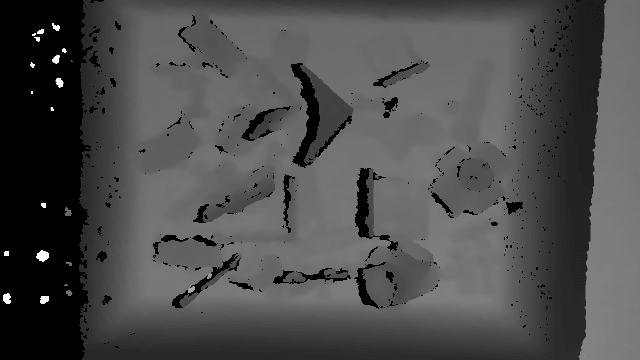}
    
    \caption{Raw depth image}
    \label{5a}
\end{subfigure}
\begin{subfigure}[t]{0.45\columnwidth}
    \centering
    \includegraphics[width=\textwidth]{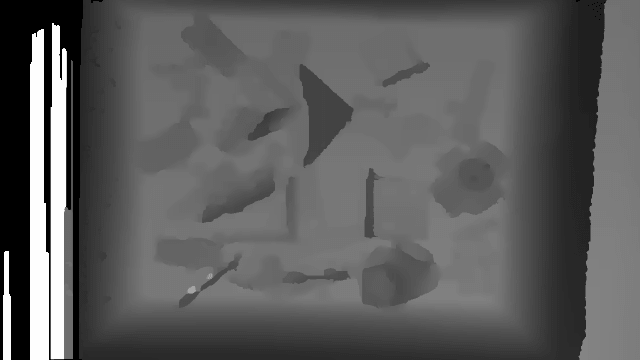}
    
    \caption{Filled depth image}
\end{subfigure}
\begin{subfigure}[t]{0.45\columnwidth}
    \centering
    \includegraphics[width=\textwidth]{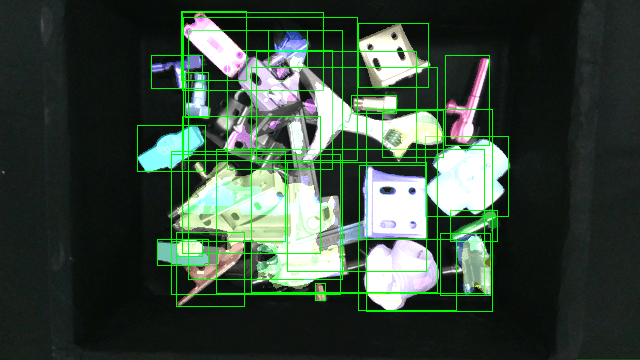}
    
    \caption{RGB only}
\end{subfigure}
\begin{subfigure}[t]{0.45\columnwidth}
    \centering
    \includegraphics[width=\textwidth]{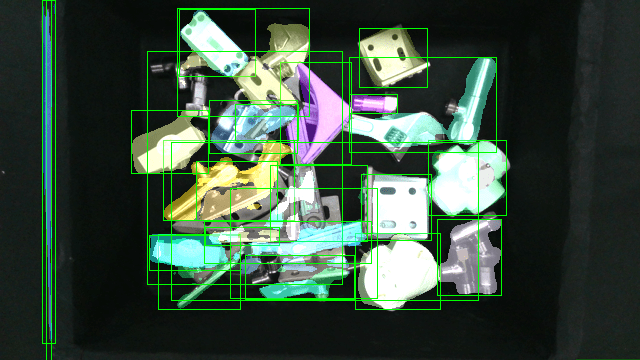}
    
    \caption{Filled depth + Early fusion}
\end{subfigure}
\begin{subfigure}[t]{0.45\columnwidth}
    \centering
    \includegraphics[width=\textwidth]{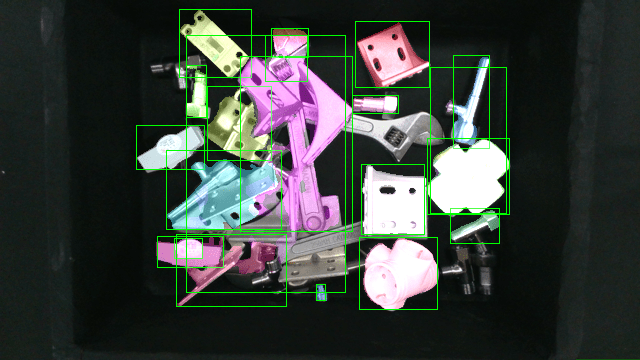}
    
    \caption{Filled depth + Late fusion}
\end{subfigure}
\begin{subfigure}[t]{0.45\columnwidth}
    \centering
    \includegraphics[width=\textwidth]{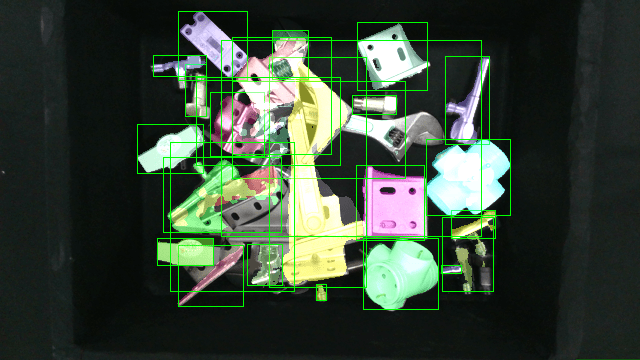}
    
    \caption{Ours}
\end{subfigure}
\caption{Input depth images (\ref{5a}, 5b) and inference results with different methods: Mask R-CNN with (5c) RGB input, (5d) RGB and filled depth using early fusion, (5e) RGB and filled depth using late fusion, and (5f) Ours.}
\label{fig:comparisons}
\end{figure}

To train the model, only synthetic data were used. Weights of the network were updated with the Adam optimizer with a learning rate of 1e-4 for 30 epochs, and the best model on validation set was tested on real data. ResNet-50 in the RGB branch was pretrained on ImageNet. For RGB images, random color jitters and Gaussian noise were applied for data augmentation. The depth cut was performed from 0.35–0.8m, and images were resized to 640x360 before being fed into the network. When filled depth maps were used as the input, missing values in the depth maps were filled with a hole filling filter with a nearest neighboring pixel closest to the sensor after applying the spatial edge-preserving filter \cite{gastal2011domain}.

The experimental results are shown in Table \ref{table:results}. We compared the performance of our model with Mask R-CNN baselines with different input modalities (RGB, filled depth, and raw depth) and RGB-D fusion strategies. When depth is applied as an input, the validity mask was connected and fed into the model with depth together. Early fusion indicates the Mask R-CNN baseline that uses the concatenation of RGB, depth maps and validity masks as inputs. The late fusion model refers to the model with RGB, a depth branch, and a fusion model but does not contain a confidence map estimator. Except for the confidence map estimator, all architecture in late fusion was the same as our proposed model.

\begin{figure}[t!]
\centering
\begin{subfigure}[t]{0.45\columnwidth}
    \centering
    \includegraphics[width=\textwidth]{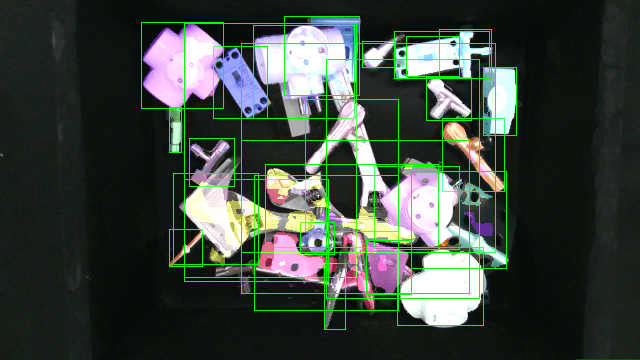}
\end{subfigure}
\begin{subfigure}[t]{0.45\columnwidth}
    \centering
    \includegraphics[width=\textwidth]{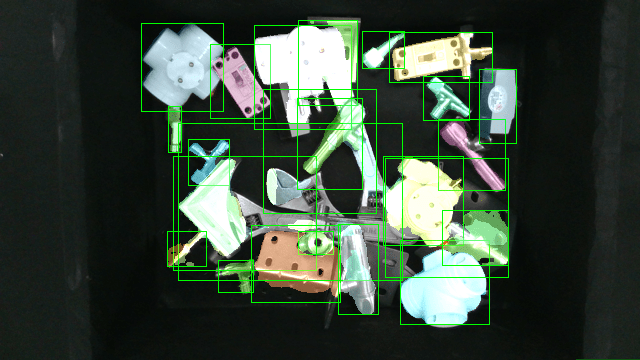}
\end{subfigure}
\begin{subfigure}[t]{0.45\columnwidth}
    \centering
    \includegraphics[width=\textwidth]{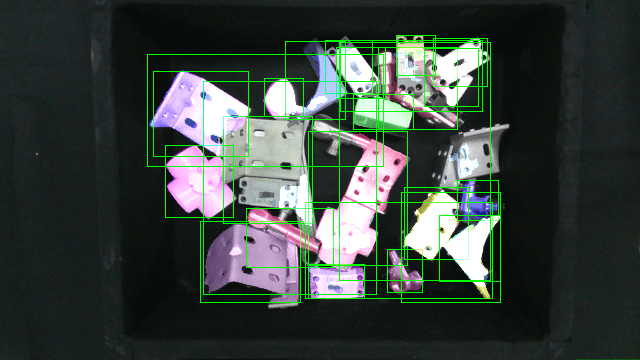}
    \caption{RGB only}
    \label{fig:rgb_confidence_rgb}
    
\end{subfigure}
\begin{subfigure}[t]{0.45\columnwidth}
    \centering
    \includegraphics[width=\textwidth]{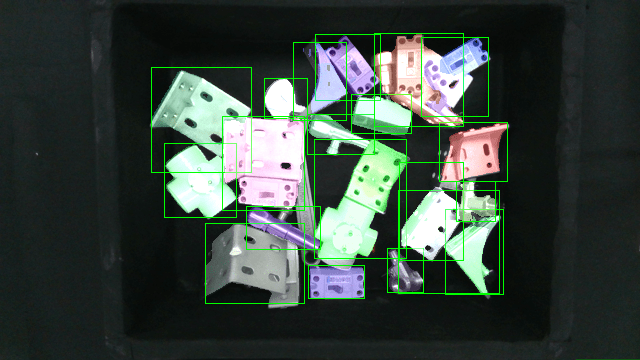}
    \caption{Ours}
    \label{fig:rgb_confidence_ours}
\end{subfigure}

\caption{Qualitative comparison between Mask R-CNN baselines with RGB inputs and our proposed model.}
\label{fig:rgb_confidence}
\end{figure}

Mask R-CNN baselines with RGB inputs show reasonable generality over unseen objects with 54.9 AP and 65.0 AR. This shows that our synthetic data generation methods can be applied to unseen industrial object segmentation without fine-tuning using real data. Among the baselines, our model with a confidence map estimator achieved the best performance compared to all others with 57.7 AP and 66.1 AR. Interestingly, the performance of the baseline that uses only RGB as an input was better than the early fusion baseline. Meanwhile, the performance of the RGB-D fusion with the raw depth model is lower than the model that uses filled depth in both early and late RGB-D fusion. This suggests that an inappropriate RGB-D fusion strategy can lower the segmentation performance, while our model can effectively exploit a raw depth map with a confidence map estimator, which leads to a significant improvement compared to other baselines. Figure \ref{fig:comparisons} and \ref{fig:rgb_confidence} show qualitative comparisons between baselines and our model, which suggest that our method can capture unseen objects more sharply than other baselines. 

\section{Conclusion}
In this study, we aimed to segment unseen industrial parts in a bin-clutter environment using only a synthetic dataset for training. We synthesized RGB-Depth-Mask pairs in simulations by randomizing object textures and mimicking real-world depth noise. To accurately segment texture-less and metallic parts, we proposed a Mask R-CNN with a confidence map estimator for RGB-D fusion, which can exploit raw depth images effectively. We showed that confidence map estimation in comparison to the baseline could lead to better generalization performance of unseen real objects, both qualitatively and quantitatively. As future work, we are planning to extend our system to unseen generic object instance segmentation and compare it with the other algorithms on public datasets.

\bibliographystyle{IEEE}
\bibliography{reference.bib}

\end{document}